\documentclass[conference, onecolumn]{IEEEtran}

\IEEEoverridecommandlockouts
\usepackage{amsmath,amssymb,amsfonts}
\usepackage{algorithmic}
\usepackage{graphicx}
\usepackage{textcomp}
\usepackage{xcolor}
\usepackage{natbib}
\usepackage{hyperref} 
\usepackage{array,booktabs}

\graphicspath{{./}} 
\def\BibTeX{{\rm B\kern-.05em{\sc i\kern-.025em b}\kern-.08em
    T\kern-.1667em\lower.7ex\hbox{E}\kern-.125emX}}

\begin{document}

\title{Model Explainability in Deep Learning Based Natural Language Processing\\
%{\footnotesize How to explain Deep Learning based models' decision in text mining?  }
}

\author{\IEEEauthorblockN{Shafie Gholizadeh}
\IEEEauthorblockA{\textit{Corporate Model Risk}\\
\textit{Wells Fargo, US}\\ 
Shafie.Gholizadeh@wellsfargo.com}
\and
\IEEEauthorblockN{Nengfeng Zhou}
\IEEEauthorblockA{\textit{Corporate Model Risk}\\
\textit{Wells Fargo, US}\\ 
nengfeng.zhou@wellsfargo.com}

}

\maketitle

  \def\mystrut(#1,#2){\vrule height #1pt depth #2pt width 0pt}    
  \def\arraystretch{1.3}%  1 is the default, change whatever you need  

\begin{abstract}
Machine learning (ML) model explainability has received growing attention, especially in the area related to model risk and regulations. In this paper, we reviewed and compared some popular ML model explainability methodologies, especially those related to Natural Language Processing (NLP) models.  We then applied one of the NLP explainability methods Layer-wise Relevance Propagation (LRP) to a NLP classification model.  We used the LRP method to derive a relevance score for each word in an instance, which is a local explainability.  The relevance scores are then aggregated together to achieve global variable importance of the model. Through the case study, we also demonstrated how to apply the local explainability method to false positive and false negative instances to discover the weakness of a NLP model.  These analysis can help us to understand NLP models better and reduce the risk due to the black-box nature of NLP models.  We also identified some common issues due to the special natures of NLP models and discussed how explainability analysis can act as a control to detect these issues after the model has been trained. 

\end{abstract}


\section{Introduction}
Machine Learning (ML) model explainability has become a vital concern in the application of ML to different fields, such as finance, medical, marketing and recommendation systems etc. When an interpretable model is available, it's preferred than a non-interpretable one when their performances are comparable due to its transparency \citep{ribeiro2016model}. However, ML models are usually not transparent and model explainability analysis can help people to understand the ML models better and trust them more. 
In the attachment of the Fed - Supervisory Letter SR 11-7 on guidance on model risk management \citep{SR117},  evaluation of conceptual soundness is one of the required core elements in effective validation framework.  ``Key assumptions and the choice of variables should be assessed, with analysis of their impact on model outputs and particular focus on any potential limitations.''  Model explainability analysis such as variable importance and feature effects methodologies can help to address the conceptual soundness of a ML model. 
In the real world, NLP models may be deployed to detect emerging trends, for example, complaints detection or communications monitoring. In such situations, past performance is not fully informative of performance for intended use case. Ongoing model explainability analysis is needed to detect the variable importance changes in the underlying models.  

Understanding the learning process in black-boxes is a challenging problem, especially in the case that a model takes hundreds or thousands of variables as input. It is hard to find out which features have the most impact on the model and how each of them is impacting the model when processing a single input record or dealing with the whole input data records in general. Here we review some methods of model explainability that can be utilized in natural language processing. All of these models are designed to find the impact of different features based on which model makes a certain decision about a single record (Local Explainability). However, the results of each method can be aggregated over all the input instances to provide the impact of features/tokens in general (Global Explainability) \citep{liu2018model}. As for a case study, we analyze the performance of Layer-wise Relevance Propagation (LRP) method for a popular text classifier—i.e., SVM working on Yelp sentiment data set.  We also did same analysis to a customer complaint data in the Bank. Because of confidentiality, we only use the Yelp example in the paper.  
 
\section{Explainability in Machine Learning, an overview}

Before we get into the special area of natural language processing, we begin with a brief review of some of the most common methods in model explainability for machine learning.

\subsection{Gradient based sensitivity analysis (GbSA)}

The simplest way to relate the inputs of a black-box to its output is to compute the partial derivative of the output with respect to each input feature. Computing the sensitivity of output to each independent variable is a computationally affordable method that inspired many more complicated (and often computationally expensive) algorithms.

\subsection{LIME (Local Interpretable Model-Agnostic Explanations) \citep{ribeiro2016model} }

In LIME algorithm, we assume the black-box model decision function might be so complicated that it cannot be approximated by a linear or simple model. To explain the model behavior for any instance, we may generate additional input instances in the local neighborhood of the original instance and ask the model to provide its decision. It is reasonable to assume the model behavior in a local neighborhood can be approximated be a simpler method like a linear model. 

\subsection{Integrated gradients (IG) \citep{sundararajan2017axiomatic} }

In Integrated Gradients method, for each observation $x$, we assume a baseline $x^{\prime}$. Integrate the partial derivative of the output with respect to each variable in the way from the baseline to the real observation. \citet{sundararajan2017axiomatic}
 suggested the following formula to compute integrated gradient, where $x_{i} $ is the $i$-th dimension on instance $x$ and $x’$ is the reference for the baseline input.

\begin{equation*}
  IG_{i}(x) \triangleq (x_{i} - x_{i}^{\prime}) \int_{\alpha=0}^{1} \frac{\partial F (x^{\prime} + \alpha \times (x-x^{\prime}))}{\partial x_{i}} d \alpha
\end{equation*}

Note that in the formula, the integration is on a curve from the reference to the instance, so the result is not trivially the integration of the first derivative.

\subsection{DeepLIFT \citep{shrikumar2017learning}}

DeepLIFT is quite similar to GbSA, but instead of taking the partial derivatives, it measures the change in model decision when a variable is changed with its reference. The results provided by DeepLIFT are possibly more stable than GbSA, but it requires a reference for each variable to compute difference function $ \Delta $.

\subsection{SHAP explanation \citep{lundberg2017unified} }

Assume that we train the proposed model on every subset $S$ of the set of all variables $(F)$. Then, including and excluding each variable i in the subset, we may get a relevance score for that variable. Finally, if we aggregate the results for each variable (weighted average of its performance in different cases), we get a relevance score for that variable. Note that SHAP can work on any model agnostic methods like LIME or model-specific methods like LRP. Let $f(x)$ denote the score function for instance $x$, then let $F$ denote the set of all input variables. SHAP explanation is driven through the formula \citep{lundberg2017unified}
:

\begin{equation*}
   \varphi_{i} =   \sum_{S\subseteq F\setminus\{i\}} \frac{|S|!(|F|-|S|-1)!}{|F|!} [f_{S\cup \{i\}}(x_{S\cup \{i\}})-f_{S}(x_{S})]
\end{equation*}

\subsection{Layer-wise Relevance Propagation (LRP) \citep{bach2015pixel} }

\begin{figure}[h]
\centering 
  \includegraphics[width=0.45\textwidth]{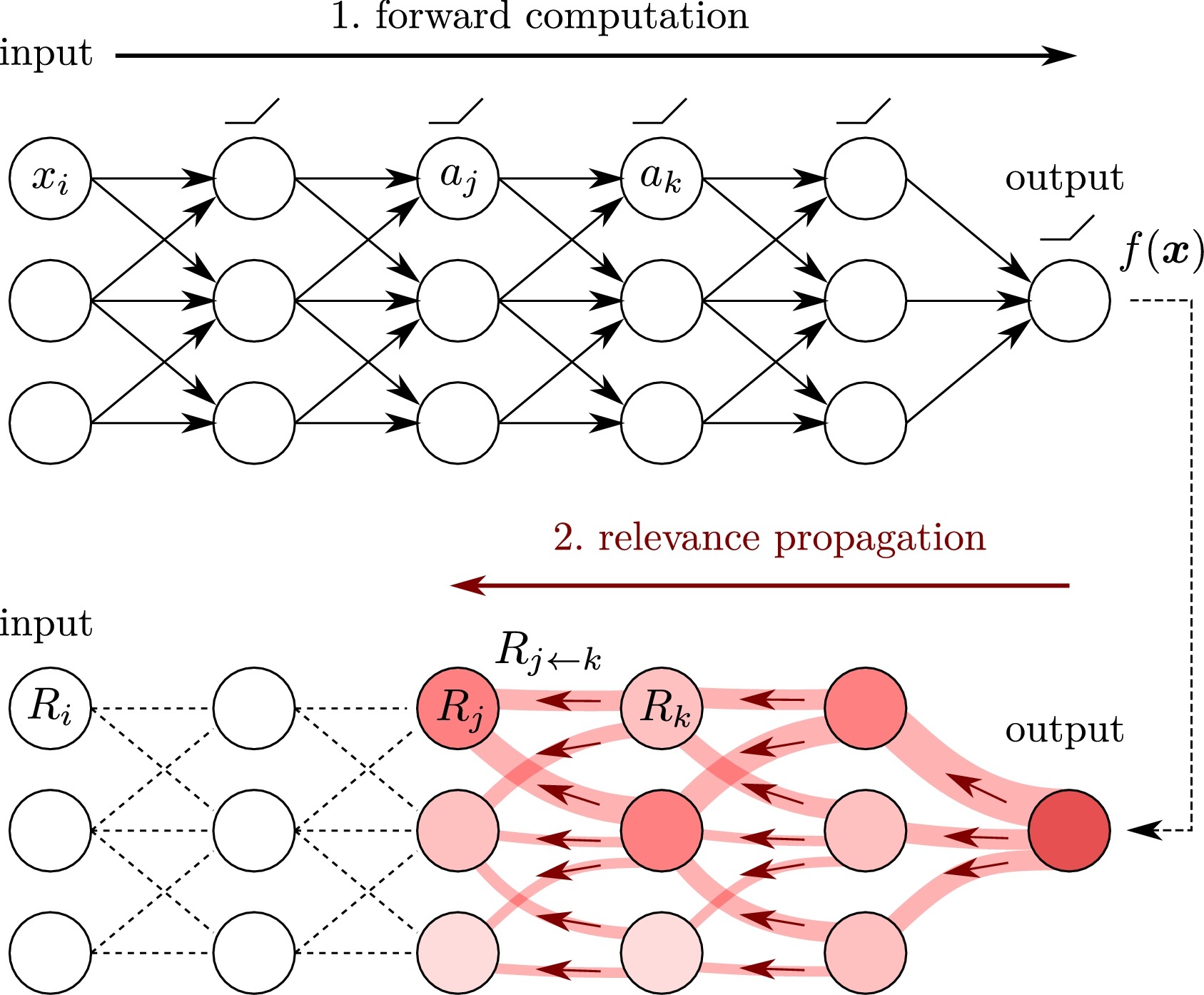}
  \caption{The flow of LRP relevance scores in the deep neural network \citep{montavon2018methods}
.}
  \label{fgr:fig1}
\end{figure}

LRP provides full decomposition of model prediction for the input to the impacts of input features. Note that LRP does not use the partial derivatives of the score function, but it decomposes the function itself. In other word, if a basic question in Sensitivity Analysis is about ``What makes this car more or less a car?'', then the basic question of LRP is ``What makes this image a car?'' \citep{arras2017relevant}
. Despite the wide range of model-agnostic explainability methods, LRP flows through the model. The flow of LRP relevance scores in a neural network is illustrated in Fig.~\ref{fgr:fig1}.

There also exist a few other explainability methods in the literature—e.g., QII (Quantitative Input Influence) \citep{datta2016algorithmic}
, LOCO (leave one covariate out) \citep{lei2018distribution}
, LIME-SUP (Locally interpretable models and effects based on supervised partitioning) \citep{hu2018locally}, additive index model for neural network \citep{vaughan2018explainable}
.

\begin{figure*}[h]
\centering 
  \includegraphics[width=0.7\textwidth]{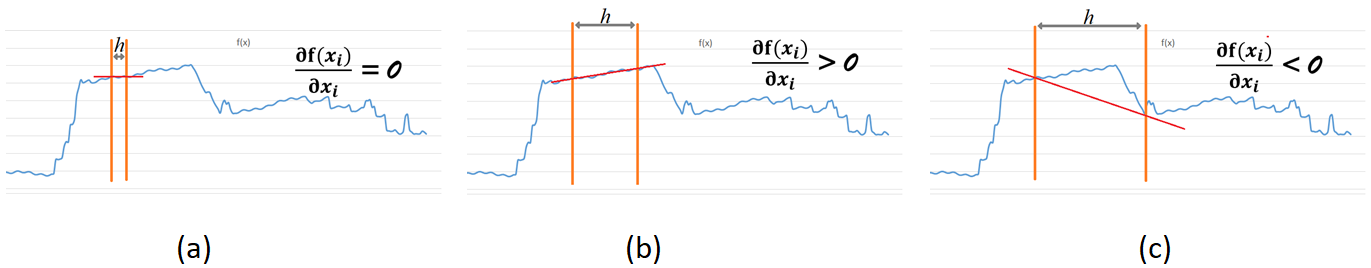}
  \caption{The impact of h value on the gradient of prediction function.}
  \label{fgr:fig2}
\end{figure*}

\begin{figure*}[h]
\centering 
  \includegraphics[width=0.85\textwidth]{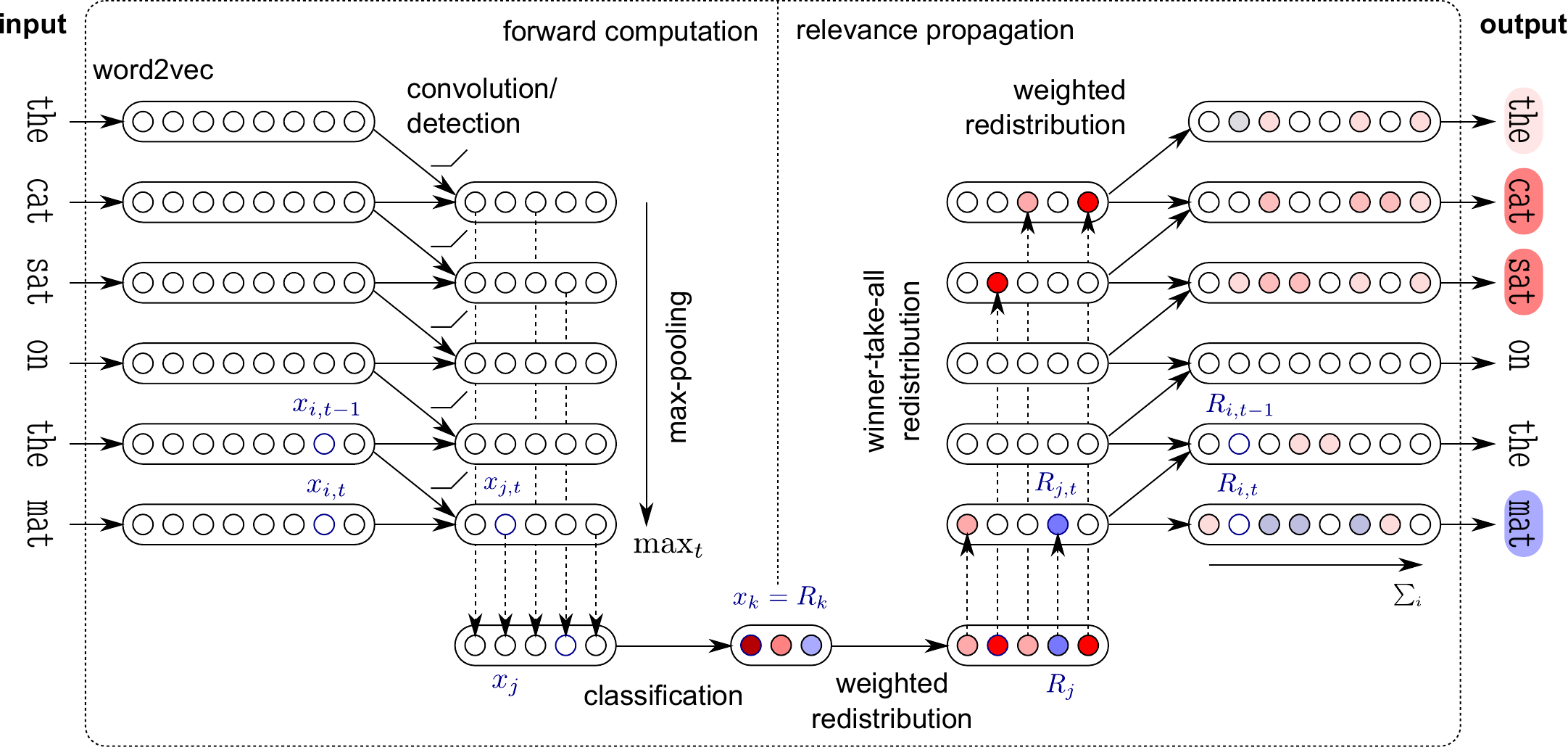}
  \caption{The schema of LRP method working on a CNN classifier \citep{arras2017relevant}
.}
  \label{fgr:fig3}
\end{figure*}

\section{Explainability in NLP}

Almost all of the methods introduced in the previous section can be used in natural language processing. However, we should note their limitations for the special case of NLP. For instance, most of the text representations in text mining come in a high number of dimensions—much higher that most numerical data sets. Also some explainability methods that fit to bag-of-word models may fail in embedding models and vice versa. Furthermore, text is set of tokens, not variables. The same indices in different textual records may refer to different entities. For instance, embedding based representation of a textual document is more like image than numerical data. Even compared to image, text could be much different and sometimes more complicated to process, e.g., in a CNN image classifier, the convolution kernel size might be 5×5 or 10×10. But for text classification it is more likely to be 3×300, 2×100 or 2×300. Furthermore, the transition from local explainability to global explainability is not easy in NLP. The easiest option is usually extracting local contributions for different tokens and then aggregating the results. It is not easily applicable when a variety of tokens/features are contributing to the model decisions. Calling for explanation, we usually demand a paragraph, a plot, or at most one page that summarizes the model performance. By aggregating all the tokens’ impacts, we obtain a global explanation in the form of an annotated vector with the size of dictionary. Note that even an absent word may be a significant determinant of the final prediction, because of its absence. Finally, the joint effect of tokens (n-grams and skip-grams impacts) are not easily distinguishable from the individual impacts.

\subsection{Gradient-Based Sensitivity Analysis }

Sensitivity Analysis can be directly or indirectly applied to textual data. Dealing with Tf-Idf continuous space, the sensitivity of the output to each input token can be directly calculated, but it does not work so easily on the embedding spaces. Assume that we need to use sensitivity analysis to the D-dimensional word embedding representation of text where each token is represented by $D$ numbers. Applying GbSA, we will get a vector of size $D$ as the sensitivity of the output to each single token. So, we need to aggregate these numbers to get only one number—the sensitivity of output to the token. As the simple summation does not make much sense for the gradient vector, the easiest way would be to use the norm of the vector. This way, sensitivity Analysis decomposes the gradient square norm of the predicting function to the impacts of the tokens. More precisely, for a token $x$, GbSA decomposes $  \|\bigtriangledown_{x}f_{c}(x) \|_{2}$ where $f_{c}$ is the prediction score on class $c$. On word embedding, this is the sum of squared partial derivatives in different embedding dimensions $d = 1, …D$. Let $x_d$ denote the $d$-th element on the embedding of token $x$. Then Let $R(x)$ denote the final relevance score of token $x$. We can easily derive \citep{arras2017relevant}
:

\begin{equation*}
   R_{d}(x) = \left( \frac{\partial f_{c}}{\partial x_{d}} (x) \right)^{2}
\end{equation*}

\begin{equation*}
  R(x) = \sum_{d} R_{d}(x) = \sum_{d} \left(  \frac{\partial f_{c}}{\partial x_{d}} (x) \right)^{2}
\end{equation*}

Then we can immediately observe that for embedding-based NLP, this relevance score does not determine polarity, unless we use some inappropriate way of aggregation, e.g., the simple summation of elements on the gradient vector as the final relevance score of the token. Note that an even more important criticism on GbSA is not limited to NLP: Why should we assume that the partial derivatives are meaningful as explanation? Sensitivity does not necessarily imply relevance, it might imply pure noise. Also, note that the relevance score is extremely sensitive to the parameter $h$—i.e., the step in numerical calculation of the partial derivative, as follows.

\begin{equation*}
    \frac{\partial f_{c}}{\partial x_{d}} (x)  = \frac{ f_{c}(x+h) -  f_{c}(x) }{h}
\end{equation*}

As shown in Fig.~\ref{fgr:fig2}, changing the $h$ value may significantly shift the results. In most applied problems, choosing the optimal $h$ is a trade-off between variance and bias, but bias is not so important in the case of GbSA. So, choosing a large $h$ may lead to the more stable results. Note that increasing $h$ is only the simplest way. The extensions of sensitivity analysis like DeepLIFT and Integrated Gradient are addressing the issue in a similar, but more reliable way, though they may require more computational cost.

\subsection{LRP \citep{arras2017relevant, arras2017explaining} }
 
The main difference between LRP and GbSA is that LRP decomposes the function itself to sum of the variables’ impacts, while Sensitivity Analysis decomposes the gradient square norm of the predicting function. 

\citet{arras2017relevant}
 utilized LRP to decompose the decision function of a CNN text classifier and used the relevance scores to provide highlighted text. Their algorithm is as follows.
\begin{enumerate}
  \item Forward-propagate numerical representations (word embeddings) through CNN.
  \item Backward-propagate output using LRP.
  \item Pool the relevance score associated to each variable/word.
\end{enumerate}

The authors suggested different back-propagation ways for different layer types:
\begin{itemize}
  \item Dense Layers: Redistribute relevance score proportional to the network weights and the inputs of the layer.
  \item Max Pooling Layers: Use the ArgMax in the previous layer and go for winner-takes-all.
  \item Convolutional Layers: Redistribute relevance score proportional to the network weights and the inputs of the layer.
  \item Input Layer: Relevance score of each word is the simple sum of its embedding elements’ scores.
\end{itemize}

\begin{figure*}[h]
\centering 
{%
\setlength{\fboxsep}{0pt}%
\setlength{\fboxrule}{1pt}%
\fbox{  \includegraphics[width=0.8\textwidth]{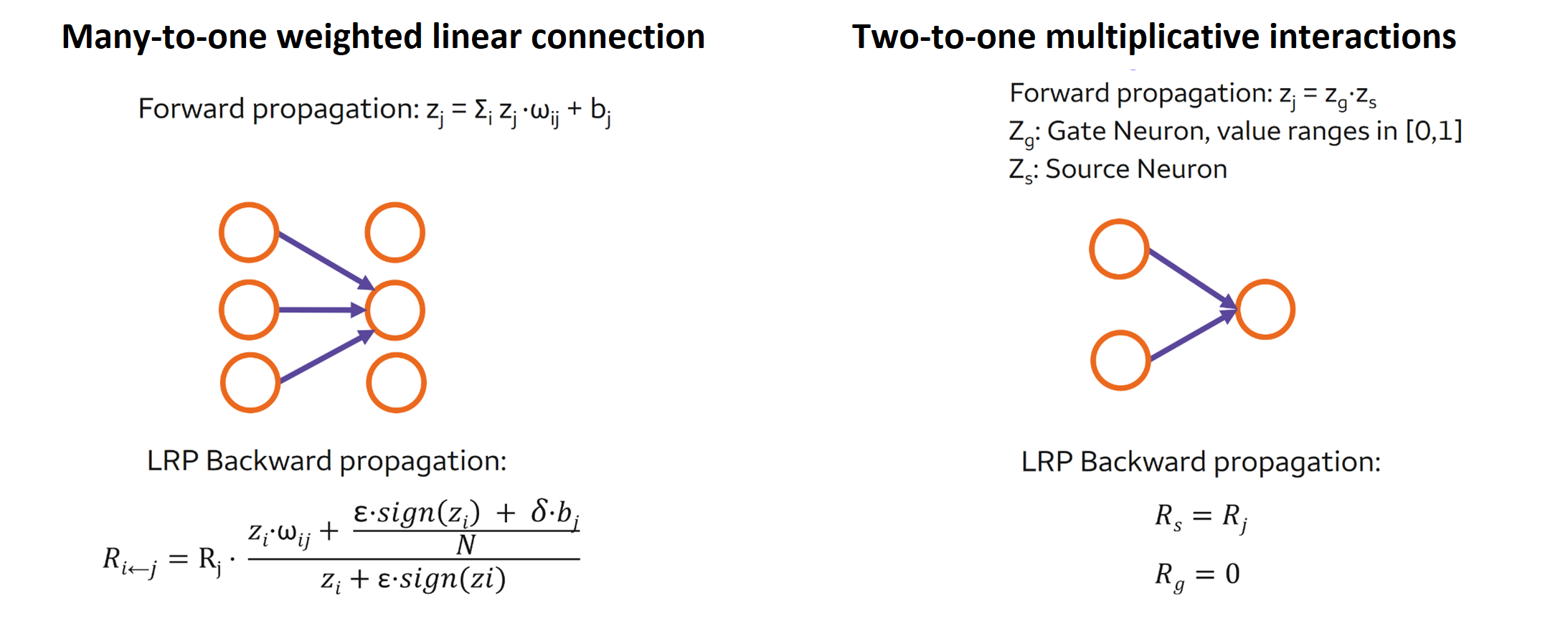}}%
}%
  \caption{Handling different types of connections in RNN based LRP \citep{arras2017explaining}
.}
  \label{fgr:fig4}
\end{figure*}

A schema of LRP method on CNN classifier is shown in Fig.~\ref{fgr:fig3}. Note that LRP is unable to distinguish the individual impacts of tokens and the joint effect of the n-grams and skip-grams. But, maybe an extra step of statistical analysis is enough to retrieve the most important joint effects. Also, the algorithm as described can only work on feed-forward networks. But  \citet{arras2017explaining}
 extended the mentioned algorithm to work on recurrent neural networks as well. The main difference is the presence of source/gate connections in RNNs, as shown in Fig.~\ref{fgr:fig4}.

\subsection{Going back from convolutional filters to ngrams }
 
\citet{jacovi2018understanding}
 examined the hypothesis that in a CNN classifier, the filters followed by max-pulling capture the fingerprint of the ngrams and extract the relevant ones respectively, then the rest of the network uses those features to classify the records. The authors retrieved the ngrams that could pass the filter’s threshold and clustered them according to their activation patterns. This way they derived a concrete identity for each filter. They showed that max-pulling is discriminating the outputs of convolutional layer—i.e., fingerprints of the relevant and non-relevant ngrams. So, their work can be utilized as an explainability method to retrieve important ngrams. Generally the idea is quite similar to the LRP method in the sense of model dependency and trying to go back through the network. However the formulation is different here, and also the final results are important ngrams.
\citet{zhao2020SHAP} used the max-pooling concept in \citet{jacovi2018understanding} to extract features in the convolutional layers and utilize SHAP to generate local explainability in text classification models.

\subsection{Self-interpretable CNN for Text Classification}
 
\citet{zhao2021SelfCNN}
 developed an approach for interpreting convolutional neural networks for text classification problems by exploiting the local-linear models inherent in ReLU-DNNs.  To get an overall self-interpretable model, the system of local linear models from the ReLU DNN are mapped back through the max-pool filter to the appropriate n-grams. The proposed technique can produce parsimonious models that are self-interpretable and have comparable performance with respect to a more complex CNN model.  This approach is using the exact coefficients of local linear models from ReLU networks.

\subsection{Other explainability methods in NLP }

% Table generated by Excel2LaTeX from sheet 'final_evaluation_iNNvestigate'
\begin{table*}[t] 
 % Table generated by Excel2LaTeX from sheet 'Sheet4' 
  \centering
  \caption{Pros and cons the different explainability methods.}
    \begin{tabular}{|p{7em}|c|p{4.945em}|c|c|c|r|}
    \hline
    Model & \multicolumn{1}{p{5.335em}|}{Distinguish Joint impacts} & \multicolumn{3}{p{13.945em}|}{Computational Cost} & \multicolumn{1}{p{4.02em}|}{Model Agnostic} & \multicolumn{1}{p{9.445em}|}{Notes} \\
    \hline
    \multicolumn{1}{|r|}{} &       & Cheap & \multicolumn{1}{p{4.945em}|}{Expensive} & \multicolumn{1}{p{4.755em}|}{Unaffordable} &       &  \\
    \hline
    GbSA  &       & X     &       &       & \multicolumn{1}{p{4.72em}|}{X} & \multicolumn{1}{p{9.445em}|}{Results might be unstable in some cases} \\
    \hline
    LRP   &       & X     &       &       &       &  \\
    \hline
    LIME  &       & \multicolumn{1}{c|}{} & \multicolumn{1}{p{4.945em}|}{X} &       & \multicolumn{1}{p{4.72em}|}{X} &  \\
    \hline
    SHAP  & \multicolumn{1}{p{5.335em}|}{X} & \multicolumn{1}{c|}{} &       & \multicolumn{1}{p{4.055em}|}{X} & \multicolumn{1}{p{4.72em}|}{X} &  \\
    \hline
    Integrated gradients &       & \multicolumn{1}{c|}{} & \multicolumn{1}{p{4.945em}|}{X} &       & \multicolumn{1}{p{4.72em}|}{X} & \multicolumn{1}{p{9.445em}|}{Requires a reference for each variable} \\
    \hline
    DeepLIFT &       & X     &       &       & \multicolumn{1}{p{4.72em}|}{X} & \multicolumn{1}{p{9.445em}|}{Requires a reference for each variable} \\
    \hline
    Map conv. filters to ngrams* &       & X     &       &       &       &  \\
    \hline
    \end{tabular}%
  \label{tab:table1survey}% 
\end{table*}

Table~\ref{tab:table1survey} shows a brief comparison of the pros and cons of some explainability methods. Here we compare only a few methods, but note that all of the methods that were mentioned for general machine learning explainability can be utilized for the specific task of text mining models and actually they have been used in the literature. Also, many other methods specifically designated for NLP models are available in the literature.  \citet{koupaee2018analyzing}
 applied visualization techniques to examine how a convolutional neural network can distinguish NLP features and how each feature is contributing to the model performance. \citet{liu2018towards}
 used fine-grained information to help explain the decision made by classification model. The framework learns to make classification decisions and generate fine-grained explanations at the same time. It uses an explainable factor and the minimum risk training approach that learn to generate more reasonable explanations. Their algorithm needs a set of golden explanations as input. \citet{raaijmakers2017investigating}
 exploited the internal neurons activation space. They used the activation on/off space as a source for KNN search in explanatory training data instances. The Authors retrieved explanatory records with the most similar neural layer activations. For validation, they used some semantic similarity between the test records and explanatory candidates. We also refer the reader to the other works \citep{montavon2018methods, futrell2019neural, reis2019explainable, wang2018explainable} that are highly related to the subject of model explainability in natural language processing.

\section{A Case Study }

\begin{table}[htbp]
  \centering
  \caption{Confusion matrix comparing the predictions of SVM model and actual labels (1 is for bad review).}
    \begin{tabular}{|c|c|c|cc|}
    \hline
          & \multicolumn{2}{p{7.67em}|}{Train Prediction} & \multicolumn{2}{p{7.67em}|}{Test Prediction} \\
    \hline
          & \multicolumn{2}{p{7.67em}|}{n=10,000} & \multicolumn{2}{p{7.67em}|}{n = 20,000} \\
    \hline
          & \multicolumn{1}{p{3.835em}|}{SVM  = 0} & \multicolumn{1}{p{3.835em}|}{SVM  = 1} & \multicolumn{1}{p{3.835em}|}{SVM  = 0} & \multicolumn{1}{p{3.835em}|}{SVM  = 1} \\
    \hline
    \multicolumn{1}{|p{4.72em}|}{Actual = 0} &       4,286  &          789  & \multicolumn{1}{c|}{      8,239 } &       1,703  \\
    \hline
    \multicolumn{1}{|p{4.72em}|}{Actual = 1} &          657  &       4,268  & \multicolumn{1}{c|}{      1,520 } &       8,538  \\
    \hline
    F1-score & \multicolumn{2}{c|}{0.86} & \multicolumn{2}{c|}{0.84} \\
    \hline
    \end{tabular}%
  \label{tab:table2svm}%
\end{table}%

% Table generated by Excel2LaTeX from sheet 'table 1'
\begin{table}[htbp]
  \centering
  \caption{Confusion matrices comparing the predictions of CNN surrogate model and the actual labels or SVM predictions.}
    \begin{tabular}{|c|c|c|}
    \hline
          & \multicolumn{2}{p{11.28em}|}{Train surrogate  } \\
    \hline
          & \multicolumn{2}{p{11.28em}|}{n=10,000} \\
    \hline
          & \multicolumn{1}{p{5.945em}|}{CNN = 0} & \multicolumn{1}{p{5.335em}|}{CNN = 1} \\
    \hline
    \multicolumn{1}{|p{4.72em}|}{Actual = 0} &                       4,385  &          690  \\
    \hline
    \multicolumn{1}{|p{4.72em}|}{Actual = 1} &                          753  &        4,172  \\
    \hline
    F1-score & \multicolumn{2}{c|}{0.86} \\ 
    \hline
    \multicolumn{1}{|p{4.72em}|}{SVM  = 0} &                       4,684  &          259  \\
    \hline
    \multicolumn{1}{|p{4.72em}|}{SVM  = 1} &                          454  &         4,603 \\
    \hline
    F1-score & \multicolumn{2}{c|}{0.93} \\
    \hline
    \end{tabular}%
  \label{tab:table3cnn}%
\end{table}%

% Table generated by Excel2LaTeX from sheet 'token_svm_cnn'
\begin{table*}[t] 
  \centering
  \caption{ Top tokens and their normalized scores retrieved by SVM permutation based approach and LRP method on surrogate CNN. }
    \begin{tabular}{|r|r|r|r||r|p{5.78em}|r|r|}
\hline  
    \multicolumn{4}{|r||}{SVM Permutation on Evaluation Data} & \multicolumn{4}{p{18.785em}|}{CNN-based LRP on Evaluation Data} \\ 
\hline  
    Rank  & Word  & SVM Score & LRP Score & Rank  & Word & \multicolumn{1}{r|}{LRP Score} & SVM Score \\ 
\hline  
    1     & mediocre & 1.00  & 0.75  & 1     & laid  & 1.00  & 0.86 \\ 
\hline  
    2     & pathetic & 0.96  & 0.77  & 2     & unfortunate & 0.96  & 0.68 \\ 
\hline  
    3     & lackluster & 0.96  & 0.96  & 3     & lackluster & 0.96  & 0.96 \\
\hline    
    4     & confusing & 0.90  & 0.41  & 4     & disappointing & 0.85  & 0.89 \\
\hline    
    5     & flavorless & 0.90  & 0.52  & 5     & crappy & 0.80  & 0.83 \\
\hline    
    6     & disappointing & 0.89  & 0.85  & 6     & outrageous & 0.77  & 0.28 \\
\hline    
    7     & poor  & 0.86  & 0.61  & 7     & pathetic & 0.77  & 0.96 \\
\hline    
    8     & worse & 0.86  & 0.61  & 8     & lacking & 0.76  & 0.77 \\
\hline  
    9     & shitty & 0.83  & 0.76  & 9     & shitty & 0.76  & 0.83 \\
\hline  
    10    & crappy & 0.83  & 0.80  & 10    & mediocre & 0.75  & 1.00 \\ 
\hline
    \end{tabular}%
  \label{tab:table4toptoken}% 
\end{table*}

As a case study on the performance of LRP methodology, we used a SVM classifier to run over real world NLP data. The data set is provided by Yelp \citep{Yelp2013} as part of a 2013 data set challenge for training and testing the prediction models.  The data set includes 229,907 reviews from Phoenix and contains information about 11,537 businesses, 43,873 users. The label of review data is customer’s rating from 1 star to 5 stars. In this case study, we will focus on binary class classification problem. We transferred the label into binary case.  We will relabel star 1 and star 2 as new label 1 (bad review).  These are more important events which can help to improve the business practice by learning from these complaints. We will relabel star 4 and star 5 as new label 0 (good review).  We did not include star 3 data in this analysis.  

In the training data we sampled 10,000 observations, using stratified random sampling to get equal number of observation for each star. There are two reasons that we did not use a very large training data in this analysis. In many real problems, the labeled data is very expensive to obtain. A training sample size of 10,000 is closer to many real problems.  The second reason for choosing small sample size is that this can help us to evaluate if the SVM model has some over fitting issues and how the over fitting can impact the model explainability when the model is applied to out of samples.

In the evaluation data we sampled 20,000 observations with same sampling method as training (actual labels available). In many other real problems, the evaluation data is usually unlabeled.  To mimic the real challenge of unlabeled data, we only need the predicted labels of the evaluation data in the NLP explainability analysis.  Note that, in real problems with only large number of unlabeled data, we can use the trained SVM model to predict the labels for unlabeled evaluation data set first.  The unlabeled data could be very unbalanced as well.  We can select some records with the highest probability of positive class and negative class to create a relative balanced data set. 

We used the 300d word embedding of these observations (provided by pre-trained vectors of ConceptNet NumberBatch v17.06) and got the average of 300 embedding dimensions of each observation to feed into the SVM model. 
Table~\ref{tab:table2svm} show the confusion matrix on the training data and evaluation data comparing the predictions of SVM model and actual label (1 is for bad review, 0  for good review). The model's F1-score was 86\% for training data and 84\% for evaluation data. The performance on evaluation data is very comparable with the training data for the SVM model.

For the next step, we tried a permutation based approach to get the importance of tokens in SVM classifier. To do this, for each observation, we removed the tokens one-by-one to see how the probability that model assigns the record to the positive class changes. The contribution of all tokens were kept in a separate list and finally used to provide the list of most influential tokens, i.e., tokens with the highest positive average impact to the positive class. We took this approach on the positive records of the training data and the positive records of the evaluation data separately.

\begin{figure}[h]
\centering 
{%
\setlength{\fboxsep}{0pt}%
\setlength{\fboxrule}{1pt}%
\fbox{  \includegraphics[width=0.45\textwidth]{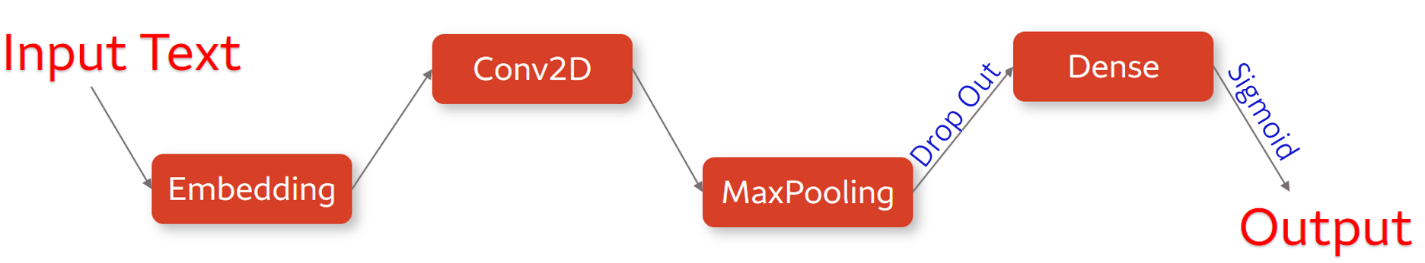}}%
}% 
  \caption{A schema of the CNN classifier.}
  \label{fgr:fig5}
\end{figure}

 \bgroup
{\setlength{\fboxsep}{ 1.5pt} \setlength{\fboxrule}{0pt}  \colorbox{white!0}   
%\usepackage{adjustbox}
%\tcbset{width=0.9\textwidth,boxrule=0pt,colback=red,arc=0pt,auto outer arc,left=0pt,right=0pt,boxsep=5pt}

\def\arraystretch{1.9}%  1 is the default, change whatever you need 
%\vspace{0.753in} 
\begin{table*}[t] 
  \caption{Training data records highlighted based on the LRP scores, where red tokens are contributing to the class 1 (bad review).}
  \centering
 \begin{tabular}{|p{0.5cm}|p{0.5cm}|p{0.5cm}|p{0.5cm}|p{14cm}|}  
%\begin{tabular}{| >{\normalsize}p{0.5cm}|>{\normalsize}p{0.5cm}|>{\normalsize}p{0.5cm}|>{\normalsize}p{0.5cm}|>{\normalsize}p{12cm}|}  
     \hline
 ID &  Star &  SVM  &  CNN  & Actual Label = 1 (Bad, Star = 1, 2 ),   SVM Pred. Label = 1 (Bad)      \\
    \hline
  1  &  1 &  0.98 &  1.0 & Getting \colorbox{red!100.0}{\strut worse}  \colorbox{red!57.0}{\strut not}  \colorbox{red!11.0}{\strut better}  1 30 appointment for a diagnostic \colorbox{blue!13.0}{\strut and}  charge the AC got the car back after 6 00pm Sent \colorbox{blue!16.0}{\strut a}  \colorbox{red!66.0}{\strut poor}  17 year old kid to \colorbox{blue!13.0}{\strut pick}  us up as their courtesy driver once it \colorbox{red!17.0}{\strut was}  finally \colorbox{blue!15.0}{\strut ready}  to go \\
    \hline
   2 & 1 &  0.99 &  1.0 & Unfortunately there is \colorbox{red!35.0}{\strut nothing}  special about this place My husband got the french dip \colorbox{blue!11.0}{\strut and}  myself the mushroom \colorbox{blue!13.0}{\strut panini}  Mine was \colorbox{red!24.0}{\strut rather}  \colorbox{red!100.0}{\strut disappointing}  the mushrooms were minced so tiny and the flavor was semi reminiscent of canned cream of mushroom soup on a \colorbox{blue!14.0}{\strut sandwich}  I hate leaving \colorbox{red!32.0}{\strut bad}  reviews but it wouldn t help anyone if i lied \colorbox{red!13.0}{\strut sorry}  \\

    \hline
 3 &     2 &  1.0 &  0.99  & Over \colorbox{blue!14.0}{\strut priced}  \colorbox{blue!18.0}{\strut and}  \colorbox{red!100.0}{\strut mediocre}  food \\
    \hline
 4 &     2 &  1.0 &  1.0  & The \colorbox{red!100.0}{\strut nasty}  youngster working at the Wetzell s Pretzel counter \colorbox{red!36.0}{\strut ruined}  it man She \colorbox{red!15.0}{\strut was}  all pissed at me because she misheard my order \colorbox{blue!35.0}{\strut and}  I \colorbox{red!53.0}{\strut bothered}  her to give me the right kind of \colorbox{blue!22.0}{\strut pretzel}  Lame Grow up little girl Rude
 \\
    \hline
 5 &     1 &  1.0 & 1.0  & Duh what \colorbox{blue!15.0}{\strut a}  wasteland of \colorbox{red!100.0}{\strut crappy}  products Gift card forced me to pop by in disguise \\
    \hline
\end{tabular}
  \label{tab:table5badbad}
\end{table*}  
} 

{\setlength{\fboxsep}{ 1.5pt} \setlength{\fboxrule}{0pt}  \colorbox{white!0}   
%\usepackage{adjustbox}
%\tcbset{width=0.9\textwidth,boxrule=0pt,colback=red,arc=0pt,auto outer arc,left=0pt,right=0pt,boxsep=5pt}

\def\arraystretch{1.9}%  1 is the default, change whatever you need 
%\vspace{0.753in} 
\begin{table*}[t] 
  \caption{False positive instances highlighted by tokens' LRP scores, where red tokens are contributing to the class 1 (bad review).}
  \centering
 \begin{tabular}{|p{0.5cm}|p{0.5cm}|p{0.5cm}|p{0.5cm}|p{14cm}|}  
%\begin{tabular}{| >{\normalsize}p{0.5cm}|>{\normalsize}p{0.5cm}|>{\normalsize}p{0.5cm}|>{\normalsize}p{0.5cm}|>{\normalsize}p{12cm}|}  
   \hline
 ID &  Star &  SVM  &  CNN  & Actual = 0 (Good, Star = 4, 5 ),   SVM Pred. Label = 1 (Bad)      \\
    \hline 
  1 &  5 &  0.99 & 0.94  &  
   I have never had \colorbox{blue!12.0}{\strut a}   \colorbox{red!39.0}{\strut bad}   \colorbox{blue!85.0}{\strut meal}   or \colorbox{red!100.0}{\strut poor}   \colorbox{red!13.0}{\strut service}   at any Ono location I really like there food and service
 \\  \hline  
  2 &  5 &  0.97 & 0.99 &  
   It s your standard Dairy Queen Delicious \colorbox{red!100.0}{\strut crappy}   for you food we all grew up on Remember getting \colorbox{blue!34.0}{\strut sundaes}   in little baseball helmets
 \\  \hline     
 3 &  4 &  0.95 & 0.89 &  
   Great food nice people \colorbox{blue!13.0}{\strut and}   \colorbox{red!100.0}{\strut mediocre}   \colorbox{red!19.0}{\strut service}   I \colorbox{blue!70.0}{\strut enjoy}   this place often
 \\  \hline  
  4 &  4 &  1.0 & 1.0 &  
   Not \colorbox{red!15.0}{\strut bad}   Had better had \colorbox{red!100.0}{\strut worse}   Will be back
 \\  \hline  
  5 &  5 &  0.59 & 0.37 & 
   Wonderful Wonderful \colorbox{blue!100.0}{\strut wonderful}   If you \colorbox{red!15.0}{\strut leave}   here less than fullfilled I think you have a food problem There is so much here you \colorbox{blue!14.0}{\strut can}   \colorbox{red!11.0}{\strut t}   \colorbox{red!19.0}{\strut possibly}   \colorbox{red!55.0}{\strut complain}   \colorbox{red!13.0}{\strut about}   \colorbox{red!44.0}{\strut not}   having enough to eat Sushi \colorbox{blue!15.0}{\strut lovers}   this is the \colorbox{blue!25.0}{\strut best}   place in town
    \\  \hline 
\end{tabular}
 \label{tab:table6badgood}
\end{table*}  
} 

{\setlength{\fboxsep}{ 1.5pt} \setlength{\fboxrule}{0pt}  \colorbox{white!0}   
%\usepackage{adjustbox}
%\tcbset{width=0.9\textwidth,boxrule=0pt,colback=red,arc=0pt,auto outer arc,left=0pt,right=0pt,boxsep=5pt}

\def\arraystretch{1.9}%  1 is the default, change whatever you need 
%\vspace{0.753in} 
\begin{table*}[t]  
  \caption{False negative instances highlighted by tokens' LRP scores, where red tokens are contributing to the class 1 (bad review).}
  \centering
 \begin{tabular}{|p{0.5cm}|p{0.5cm}|p{0.5cm}|p{0.5cm}|p{14cm}|}  
%\begin{tabular}{| >{\normalsize}p{0.5cm}|>{\normalsize}p{0.5cm}|>{\normalsize}p{0.5cm}|>{\normalsize}p{0.5cm}|>{\normalsize}p{12cm}|}  
  
 \hline
 ID &  Star &  SVM  &  CNN  & Actual = 1 (Bad, Star = 1, 2 ),   SVM Pred. Label = 0 (Good)    \\  \hline   
 1 &   2 &  0.32 & 0.7 &  
   It was alright If you need \colorbox{red!13.0}{\strut something}  quick and close it will do in a pinch Agree with others that the lack of sauce is \colorbox{red!100.0}{\strut disappointing}  Egg roll was like any other Nothing too special The steamed veggies were \colorbox{blue!39.0}{\strut pretty}  \colorbox{blue!40.0}{\strut fresh}  \colorbox{blue!16.0}{\strut and}  crisp so that was a plus
 \\  \hline    
 2 &   2 &  0.44 & 0.19 &  
   Not \colorbox{red!51.0}{\strut bad}  but \colorbox{red!72.0}{\strut not}  \colorbox{blue!79.0}{\strut great}  Service was \colorbox{blue!20.0}{\strut really}  \colorbox{blue!100.0}{\strut friendly}  \colorbox{blue!22.0}{\strut and}  prompt Food \colorbox{red!14.0}{\strut was}  meh \colorbox{red!36.0}{\strut maybe}  I just \colorbox{red!16.0}{\strut ordered}  the wrong thing
 \\  \hline    
  3 &   2 &  0.25 & 0.3 &  
   I ve \colorbox{red!20.0}{\strut tried}  time and time again to \colorbox{red!13.0}{\strut like}  this place but the pizza \colorbox{red!17.0}{\strut hmmm}  is \colorbox{red!100.0}{\strut mediocre}  at \colorbox{blue!17.0}{\strut best}  the pazookie ice cream \colorbox{blue!13.0}{\strut cookie}  is \colorbox{blue!95.0}{\strut great}  and so is the atmosphere
 \\  \hline  
  4 &   2 &  0.16 & 0.06  & 
   Fell into the it \colorbox{red!13.0}{\strut was}  on Food Network so let s \colorbox{blue!11.0}{\strut visit}  trap It \colorbox{red!11.0}{\strut was}  an okay breakfast but the service was pretty \colorbox{red!100.0}{\strut lackluster}  Not \colorbox{red!15.0}{\strut worth}  driving from Phoenix for the food however the drive itself is \colorbox{blue!19.0}{\strut very}  nice and they \colorbox{blue!12.0}{\strut have}  \colorbox{blue!18.0}{\strut a}  \colorbox{blue!90.0}{\strut lovely}  \colorbox{blue!16.0}{\strut patio}  on which you \colorbox{blue!24.0}{\strut can}  eat your eggs
 \\  \hline     
  5 &   2 &  0.14 & 0.1 & 
   If all natural means dry and bland then this \colorbox{blue!15.0}{\strut cupcakes}  \colorbox{blue!11.0}{\strut are}  \colorbox{blue!57.0}{\strut definitely}  \colorbox{blue!12.0}{\strut natural}  I was \colorbox{blue!22.0}{\strut super}  excited to try Lulus and then \colorbox{red!16.0}{\strut was}  quickly \colorbox{red!100.0}{\strut disappointed}  They \colorbox{red!25.0}{\strut were}  out of everything and the cake texture was \colorbox{red!11.0}{\strut like}  cardboard Whipped frosting \colorbox{red!14.0}{\strut ummm}  where is my cream cheese Save your 3 and go grab a \colorbox{blue!49.0}{\strut happy}  hour beer
 \\  \hline     
\end{tabular} 
 \label{tab:table7goodbad}
\end{table*}  
}
\egroup

\begin{figure*}
    \centering
    \begin{minipage}{0.5\textwidth}
        \centering
        \includegraphics[width=1\textwidth]{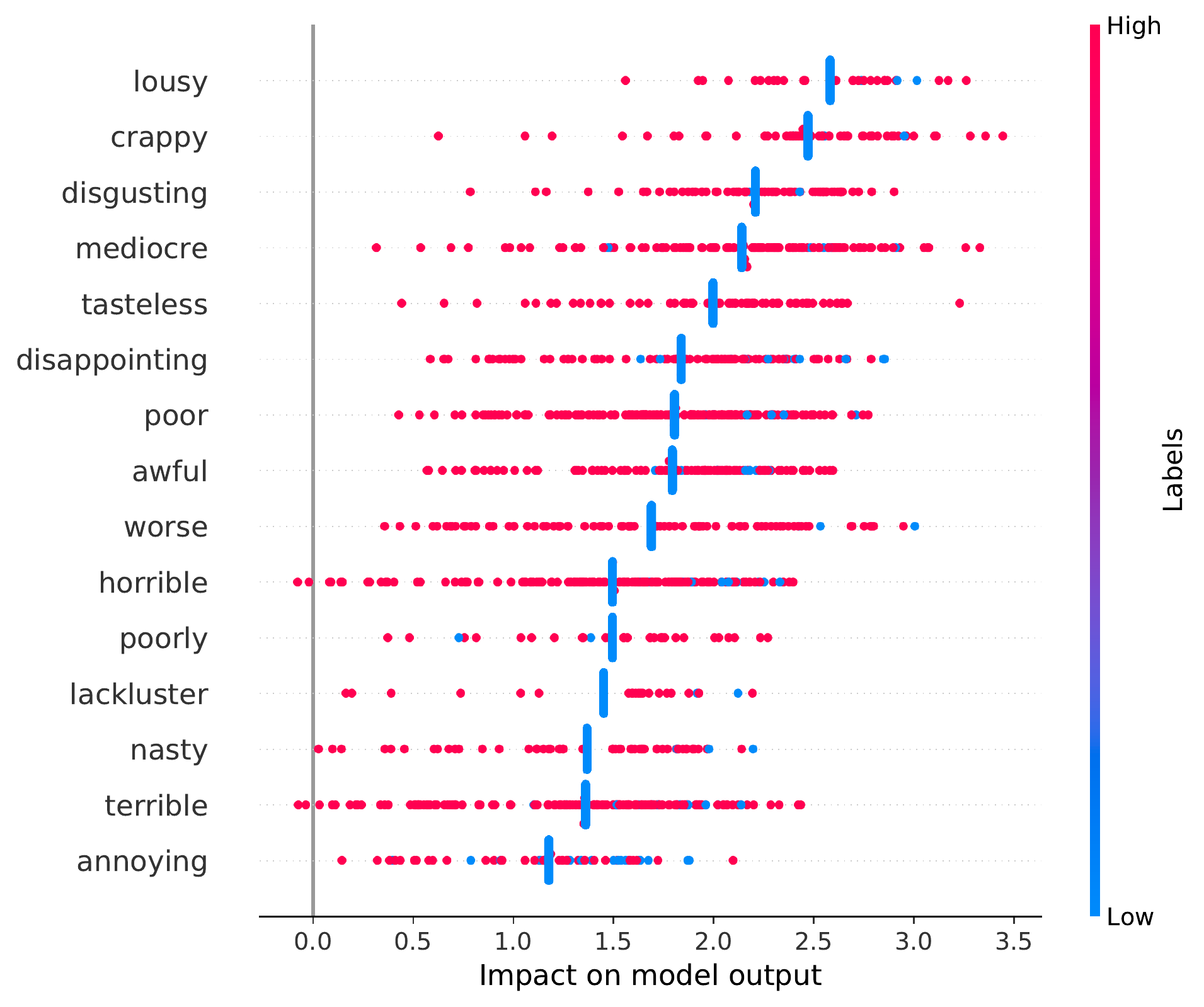}  
    \end{minipage}\hfill
    \begin{minipage}{0.5\textwidth}
        \centering
        \includegraphics[width=1\textwidth]{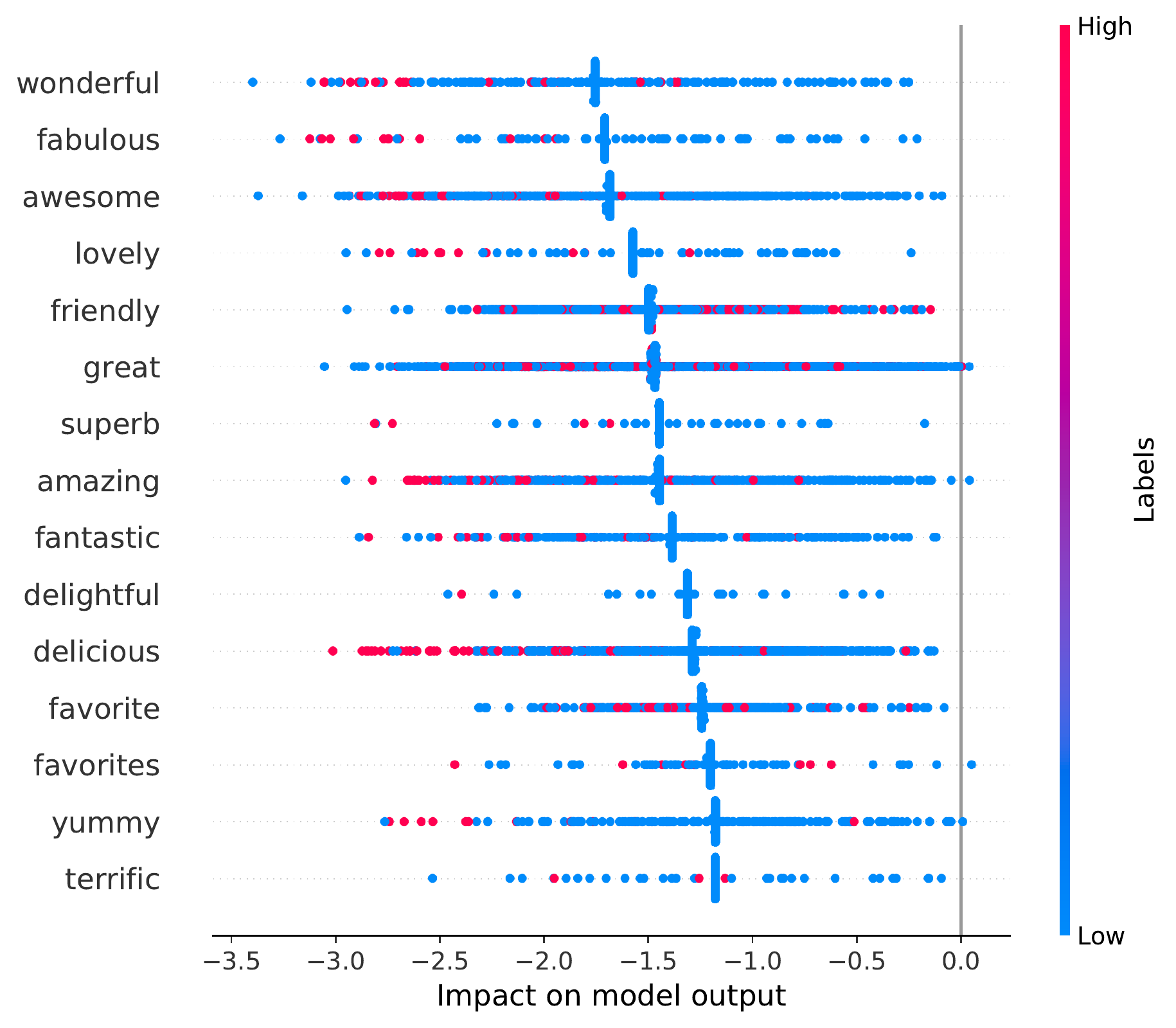}  
    \end{minipage}
        \caption{Top unigram contributing to bad review (left) and good review (right), where blue dots represent predicted good reviews by SVM model. }
  \label{fgr:fig6ngram1}
\end{figure*}
 
\begin{figure*}
    \centering
    \begin{minipage}{0.5\textwidth}
        \centering
        \includegraphics[width=1\textwidth]{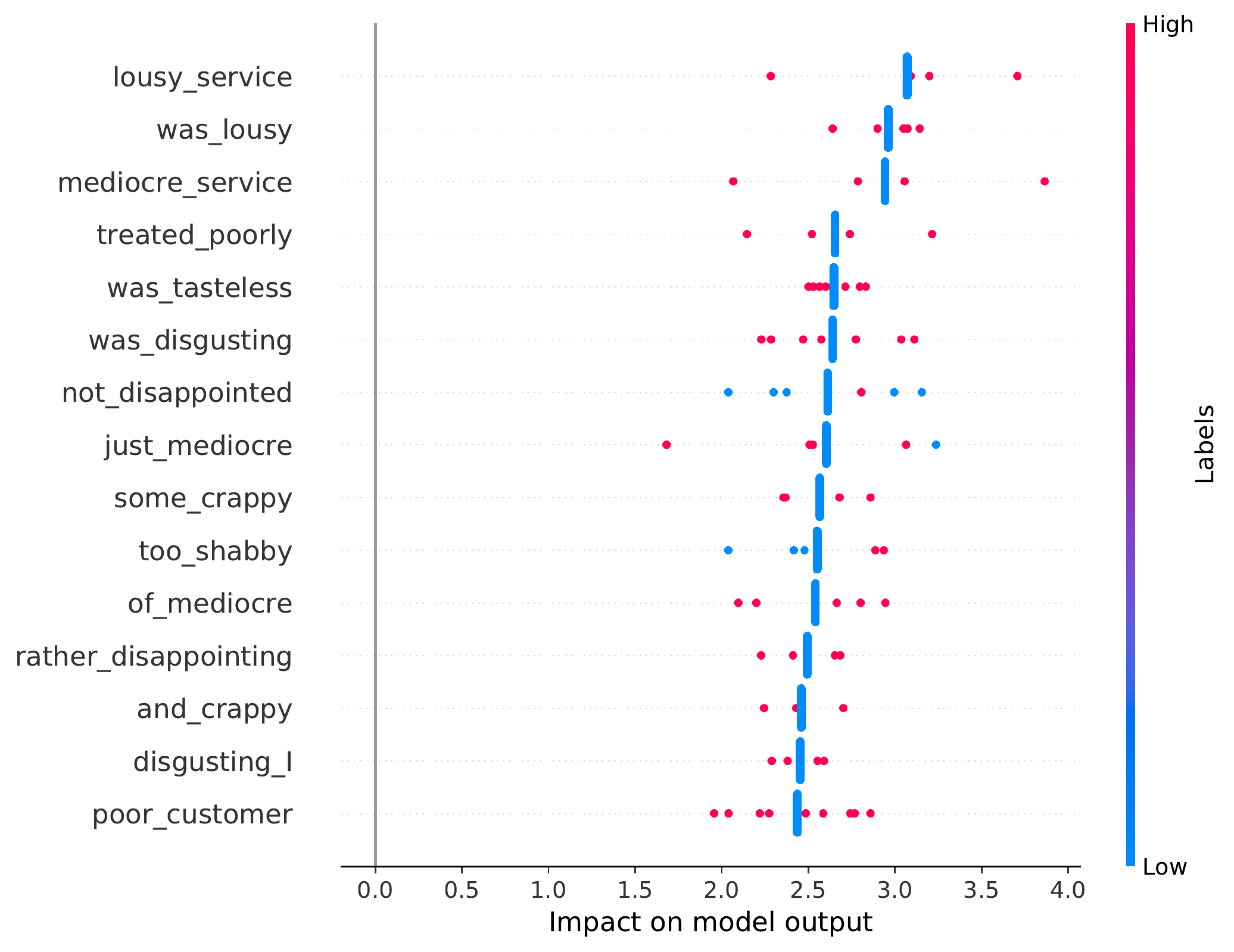}  
    \end{minipage}\hfill
    \begin{minipage}{0.5\textwidth}
        \centering
        \includegraphics[width=1\textwidth]{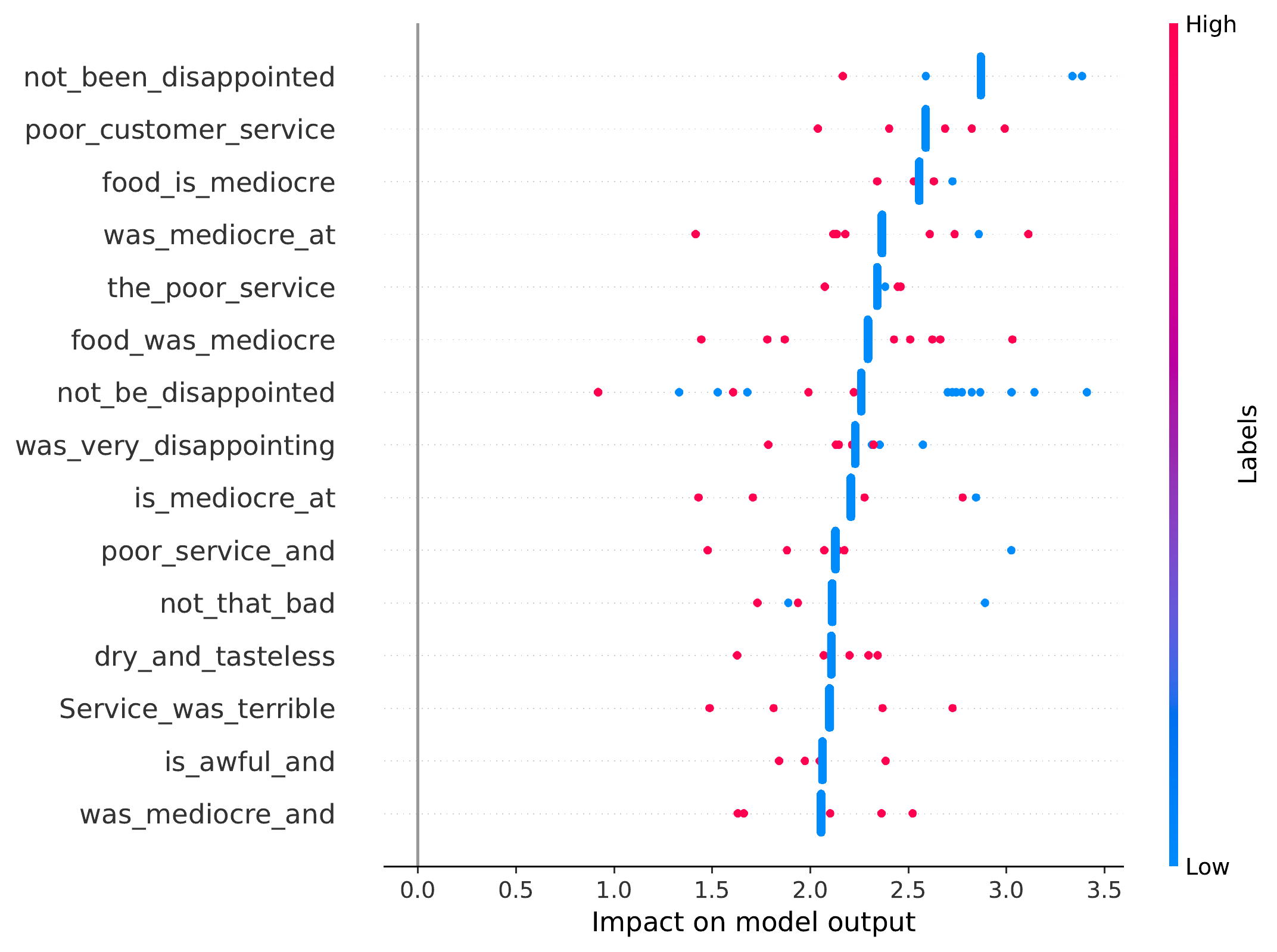}  
    \end{minipage}
        \caption{Top bigrams (left) and trigrams (right) contributing to bad review prediction.}
  \label{fgr:fig7ngram23} 
\end{figure*}

For the LRP methodology, we used a surrogate CNN model shown in Fig.~\ref{fgr:fig5}. We trained the CNN model on all of the records on the training and evaluation data set where the labels were those predicted by SVM. Table~\ref{tab:table3cnn} show the confusion matrices comparing the predictions of CNN surrogate model and the actual labels or SVM predictions.  The CNN surrogate model predicted the actual label pretty well, even it's trained using the SVM prediction as the dependent variable.   The CNN surrogate model and the SVM prediction are very consistent, with a F1 score of 0.93. 

Then we used the python package iNNvestigate \citep{alber2019innvestigation}
 to run LRP on both positive records of the training data and positive records of the evaluation data separately. The list of the most influential tokens retrieved by permutation approach and CNN-based LRP are shown in Table~\ref{tab:table4toptoken} (based on the evaluation data). For simplicity of comparison the corresponding score in the rival method is shown to the right of each model score.  Note that to filter the rare tokens, we used a threshold of 20. Also, a few training records are shown in Table~\ref{tab:table5badbad}, highlighted by their tokens’ LRP relevance scores.  The highlights help us to see clearly why the model predict these reviews as bad ones. 
  
Note that the python package ‘iNNvestigate’  contains the implementations of LRP, sensitivity analysis, etc.  Many of these methods generated very similar results. For LRP method, it works only on Tensorflow Keras-backend. It does not support networks including SoftMax activation, since SoftMax is not invertible.

A schema of the CNN classifier is shown in Fig.~\ref{fgr:fig5}. In the CNN model, for the embedding layer, the same pre-trained word embedding as SVM model (300-D ConceptNet NumberBatch 17.06) is used with the padding size (maximum number of tokens in text to be used) of 100. The embedding layer is followed by convolutional filters of sizes 2×300, 3×300, 4×300 (150 filters of each size). We used 40\% dropout after the max puling. For model training, the batch size was equal to 30 and the model was trained in five epochs.

\subsection{The special cases of False Negatives and False Positives} 

Model explainability methods enable us to examine the source of mistakes made by the model. For NLP classification models, two particular sources of mistakes are lack of enough usable tokens in the input record and model’s naïve trust on particular tokens. A few false and false negative instances in CNN surrogate model are shown in Table~\ref{tab:table6badgood} and Table~\ref{tab:table7goodbad} respectively.

The  false positives (FP) are good reviews which are predicted as bad reviews by SVM model. Some of these reviews are short and the model cannot find enough interpretable tokens and/or ngrams and then tries to decide based on whatever that is available. Some other FP errors are related to combination of negative word with negative sentiment words, such as ``never had a bad meal or poor service'' in example 1 and ``can t possibly complain'' in example 5 in Table~\ref{tab:table6badgood}. Negative word is usually related to bad review by itself. The combination of negative word with negative sentiment words is  a challenge task to learn for NLP classifications, especially when the training samples are not very large. The model may get confused dealing with the individual role of a token and its joint impact. When the sample size is not large enough, the joint impact of the bigrams or trigrams may be not strong enough to offset the combination of individual impacts. And the joint impact estimates of these negative combination words may also be weakened by other positive ngrams in the same review. 

The false negatives (FN) in Table~\ref{tab:table7goodbad} are bad reviews which are predicted as good reviews by SVM model. Most of the errors are caused by positive comment in the review regarding some specific part of the meal or service, such as ``The steamed veggies were pretty fresh and crisp'' in example 1 and ``Service was really friendly'' in example 2. And most of these reviews are star 2, which means they are bad overall with something positive.  CNN surrogate model’s prediction are consistent with SVM for most of these cases.

\subsection{Quantitative methods comparison at global level} 
 
We can add the effect tokens in ngrams to get into the list of the most influential ngrams for the model. Fig.~\ref{fgr:fig6ngram1} shows the most influential unigram contributing to bad review (left) and good review (right). Here the horizontal axis represent the impact of a unigram on the model. The blue dots represent the records whose SVM model prediction are 0 (good review). The red dots represent the records whose SVM model prediction are 1 (bad review). For the reviews which contains top tokens contributing to bad reviews the SVM model usually predicting them as bad. This shows that each of these tokens is very predictive by itself for bad reviews. The conclusion is similar for the unigrams contributing to good reviews. Since bad reviews are usually more interesting for the business to improve in the future, we focus on the analysis of bad review tokens in the case study. 

Fig.~\ref{fgr:fig7ngram23} shows the most influential bigrams and trigrams based on LRP scores, respectively. Compared to the top unigram tokens, the appearance frequencies of bigrams and trigrams are much less lower. The frequencies of reviews with the ngrams can be seen from the number of red/blue dots in corresponding to the row of the ngram. The observations without the ngram is replaced with the mean contribution of the non-missing ngrams, shown as the little bar in the middle of each row. Because of their lower frequencies, individual bigrams and trigrams impact much less number of reviews, compared to unigrams. So individually they do not have much impact for global variable importance. As we discussed earlier in the combination of negative words case in the false positive examples, the joint impact of the bigrams or trigrams are usually not strong enough to offset the individual unigram impact. This can be seen in the case of ``not disappointed'', ``not been disappointed'' and ``not be disappointed''. These ngrams belong to the top ngrams contributing to the bad reviews, due to the individual unigram impact. Most of these reviews are still predicted as good review in the SVM model, possibly due to other ngrams (contributing to good reviews) in the same reviews.

% Table generated by Excel2LaTeX from sheet 'final_evaluation_iNNvestigate'
\begin{table*}[t]
%\begin{table}[htbp]
  \centering
  \caption{Decrease in recall of the SVM predicting bad reviews by excluding top n contributing tokens returned.}
    \begin{tabular}{|c||c|c|c||c|c|c|}
    \hline
    \multicolumn{1}{|p{4.945em}||}{Number of removed tokens} & \multicolumn{1}{p{4.945em}|}{SVM permutation train data} & \multicolumn{1}{p{4.945em}|}{CNN LRP train data} & \multicolumn{1}{p{4.945em}||}{CNN GbSA train data} & \multicolumn{1}{p{5.945em}|}{SVM permutation eval. data} & \multicolumn{1}{p{4.945em}|}{CNN LRP eval. data} & \multicolumn{1}{p{4.945em}|}{CNN GbSA eval. data} \\
    \hline
    0     & 0     & 0     & 0     & 0     & 0     & 0 \\
    \hline
    50    & 0.206 & 0.285 & 0.211 & 0.209 & 0.275 & 0.212 \\
    \hline
    100   & 0.325 & 0.353 & 0.209 & 0.325 & 0.330 & 0.215 \\
    \hline
    150   & 0.389 & 0.485 & 0.273 & 0.400 & 0.472 & 0.269 \\
    \hline
    200   & 0.462 & 0.583 & 0.277 & 0.456 & 0.574 & 0.269 \\
    \hline
    250   & 0.550 & 0.655 & 0.293 & 0.535 & 0.650 & 0.291 \\
    \hline
    300   & 0.622 & 0.711 & 0.381 & 0.632 & 0.700 & 0.390 \\
    \hline
    \end{tabular}%
  \label{tab:table8recall}%
%\end{table}%
\end{table*}

\begin{figure}[h]
\centering 
   \includegraphics[width=0.45\textwidth]{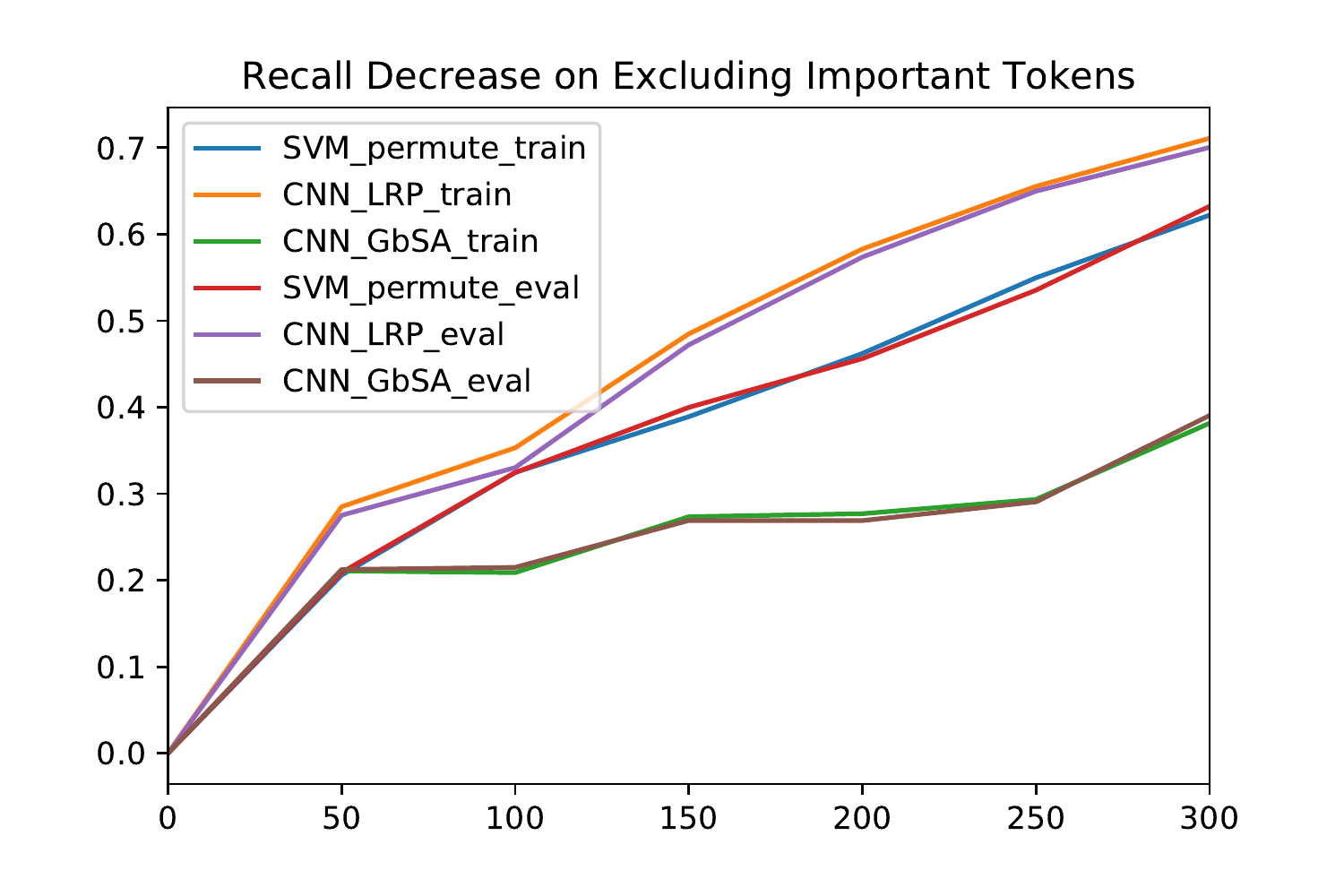}   
  \caption{ Decrease in recall of the SVM predicting bad reviews of the evaluation data.}
  \label{fgr:fig8recall}
\end{figure}

% Table generated by Excel2LaTeX from sheet 'final_evaluation_iNNvestigate'
\begin{table*}[t] 
  \centering
  \caption{Correlation matrix of the scores.}
    \begin{tabular}{|r|c|c|c|c|c|c|}
    \hline
    \multicolumn{1}{|p{6.945em}|}{Correlations of Scores} & \multicolumn{1}{p{4.665em}|}{SVM perm train} & \multicolumn{1}{p{4.555em}|}{SVM perm. eval.} & \multicolumn{1}{p{4.055em}|}{CNN LRP train} & \multicolumn{1}{p{4.055em}|}{CNN LRP eval.} & \multicolumn{1}{p{4.855em}|}{CNN GbSA train} & \multicolumn{1}{p{4.855em}|}{CNN GbSA eval.} \\
    \hline
    SVM perm train & 1.00  & 0.96  & 0.63  & 0.66  & 0.47  & 0.48 \\
    \hline
    SVM perm. eval. & 0.96  & 1.00  & 0.64  & 0.67  & 0.47  & 0.49 \\
    \hline
    CNN LRP train & 0.63  & 0.64  & 1.00  & 0.90  & 0.31  & 0.32 \\
    \hline
    CNN LRP eval. & 0.66  & 0.67  & 0.90  & 1.00  & 0.33  & 0.36 \\
    \hline
    CNN GbSA train & 0.47  & 0.47  & 0.31  & 0.33  & 1.00  & 0.95 \\
    \hline
    CNN GbSA eval. & 0.48  & 0.49  & 0.32  & 0.36  & 0.95  & 1.00 \\
    \hline
    \end{tabular}%
  \label{tab:table9cor}%
\end{table*}

Table~\ref{tab:table4toptoken} clearly provide some global explanations for the SVM model, based on two different methods. To do a comparison between the two explainability methods, we asked the SVM models to predict the positive records of evaluation data once again, after excluding top n relevant tokens returned in each of the four cases (Permutation method/LRP methods on train/evaluation data). We tried it for n=50, 100, 150, 200, 250, 300. To have a more meaningful comparison, we also utilized the same python package (iNNvestigate) to run GbSA as well, and calculated the relevance scores. Then we aggregated the scores to get the most contributing tokens (i.e. tokens with the highest average contribution) in the training data and the evaluation data found by sensitivity analysis.

The comparative results are shown in Table~\ref{tab:table8recall}. The comparison is based on the false negative ratio when we exclude top tokens.  Note that in each case, the change is measured when model is re-predicting the labels for the evaluation data. The top tokens are calculated from both the training and evaluation data. For example, in the first case, we exclude top contributing tokens found in the training data and measure the performance change in the evaluation data.  

Fig.~\ref{fgr:fig8recall} shows the comparison visually.  It can be seen that removing LRP top tokens will decrease the recall rate the most. For the same method, the top tokens from training data and evaluation data have very similar performance impact. This is related to their high score correlation as shown in Table~\ref{tab:table9cor}.

The correlation among different scores are shown in Table~\ref{tab:table9cor}. To remove noise in the correlation calculation, only the words with at least 20 appearances in evaluation data are used.  We can see that the highest correlation is between training and evaluation method for the same methodology, which explained their similar performance impacts in Fig.~\ref{fgr:fig8recall}.  There are some correlations between the SVM permutation method and LRP method.  The GbSA method has low correlations with the other two methods.

\section{COMMON ISSUES TO WATCH OUT FOR NLP MODEL}

Due to the special natures of NLP models, the followings are some common issues to watch out.  The explainability analysis can act as a control to detect these issues after the model has been trained.

a. Out-of-Vocabulary Words (OOV), Non English words or spelling errors. For NLP models using embedding, OOV are those words which cannot be found in the word embedding mapping file. Without proper handing of the missing values, the model's performance will be deteriorated and the model prediction for observations with OOV may be poor or biased. In models using English embedding, Non English words will become OOV and they are very likely to appeared together in the same observations. This could lead to very poor performance for these Non English observations. Spelling errors belong to another source of OOV, which need be be checked and fixed. Model's performance for observations with OOV need to be measured separately to assess the impact of OOV.  

b. Proxies for contextual information – e.g. words from standard disclaimers.  These proxies should be removed in the data pre-processing before modeling.  After a model has been created, checking if these proxies are in top features should be done as part of the variable importance analysis. 

c. Proper nouns – e.g. name of CEO.  When proper nouns appeared in top features with strong predictive power, assessment is needed to make sure they are reasonable. If the connection between a proper noun and dependent variable cannot be justified or the relationship is likely to change in future, these words should be removed in the data and the model need to be retrained.   

d. Handling of Negation, Sarcasm, Idioms.  NLP model may not be able to handle these situations well, as we discussed in some of the false positive examples.  Large number of observations are needed for a NLP model to learn these special situations.  Some transfer learning methodologies may benefit a NLP model to learn these words/phrases. 
 
e. Asymmetric treatment of demographic proxies.  The unstructured data used in NLP model may contain ``digit footprint'', which can be predictive for some sensitive variables such as gender, race or age etc.  This can lead to the asymmetric treatment for different classes of people.   It will cause fairness concern if the model's predictions are unfavorable for the protected class.  
 
f. Quality of the labels in NLP models.  The labels used in the NLP model are usually created by people, either by those produced the data or the some other reviewers.  The labels may be not consistent and are subjective to human judgment errors.  This can impact the model's prediction power, and cause some of the false positive/negative errors.  Some assessment of the label quality is needed as part of the model validation or review process.  
 
\section{Declaration of Interest}

The authors report no conflicts of interest. The authors alone are responsible for the content and writing of the paper.

\section{Conclusion}

As we discussed in the overview section and compared the explainability methods in Table~\ref{tab:table1survey}, Layer-wise Relevance Propagation is a reliable and still a computationally affordable explainability method. In our case study, we showed that explainability provided by LRP methodology outperform the gradient only based and permutation based explainability. By doing the NLP explainability analysis, we can get a lot insight into the black-box NLP models and reduce the risk of using a wrong or inappropriate model. The false positive and false negative examples highlighted by LRP scores also helped us to understand why the model made wrong predictions. This can help us to improve the model in the future.
 
\section{Acknowledgment}
   
We thank Harsh Singhal for useful discussion. We thank corporate risk - model risk at Wells Fargo for support.

\bibliography{./main}
\bibliographystyle{apalike}

\end{document}